\documentclass[10pt, letter, onecolumn]{arxiv}

\usepackage{adjustbox}
\usepackage{booktabs}
\usepackage{siunitx} 
\usepackage{longtable}
\usepackage{xcolor}
\usepackage{colortbl}
\usepackage{caption}
\usepackage{subcaption} 
\usepackage{hyperref}
\usepackage{geometry}
\usepackage{ragged2e}
\usepackage{makecell, tabularx}
\newcolumntype{L}{>{\RaggedRight}X}
\usepackage{siunitx}
\usepackage{comment}
\usepackage{multirow}
\usepackage{datatool}
\newcommand{\grayrow}{\rowcolor[gray]{.9}}
\usepackage{xcolor}
\colorlet{darkgreen}{green!65!black}
\colorlet{darkblue}{blue!75!black}
\colorlet{darkred}{red!80!black}

\colorlet{darkgreen}{green!65!black}
\colorlet{darkred}{red!80!black}

\usepackage[symbol]{footmisc}

\theoremstyle{thmstyleone}%
\theoremstyle{thmstyletwo}%
\theoremstyle{thmstylethree}%

\title{{\LARGE{Estimating Blood Pressure with a Camera:\\ An Exploratory Study of Ambulatory Patients with Cardiovascular Disease}}}

\author[$\ast$, 1]{Theodore Curran} 
\author[$\ast$, 2]{Chengqian Ma} 
\author[$\ast$, 2]{Xin Liu} 
\author[$\ast$, 2]{Daniel McDuff} 
\author[2]{\\Girish Narayanswamy} 
\author[3]{George Stergiou}
\author[2]{Shwetak Patel}  
\author[1]{Eugene Yang $\ddagger$}  

\affil[1]{Division of Cardiology, University of Washington School of Medicine, }
\affil[2]{Paul G. Allen School of Computer Science \& Engineering, }
\affil[3]{National and Kapodistrian University of Athens, School of Medicine}

\renewcommand{\correspondingauthor}[1]{$\ast$~Equal contributions: \{thcurran@health.ucsd.edu; cm74@uw.edu; xliu0@cs.washington.edu; dmcduff@cs.washington.edu\}  \\%
                                       $\ddagger$~Corresponding authors: \{eyang01@uw.edu\}}

\begin{document}


\begin{abstract}
Hypertension is a leading cause of morbidity and mortality worldwide. The ability to diagnose and treat hypertension in the ambulatory population is hindered by limited access and poor adherence to current methods of monitoring blood pressure (BP), specifically, cuff-based devices. Remote photoplethysmography (rPPG) evaluates an individual's pulse waveform through a standard camera without physical contact.  Cameras are readily available to the majority of the global population via embedded technologies such as smartphones, thus rPPG is a scalable and promising non-invasive method of BP monitoring. The few studies investigating rPPG for BP measurement have excluded high-risk populations, including those with cardiovascular disease (CVD) or its risk factors, as well as subjects in active cardiac arrhythmia. The impact of arrhythmia, like atrial fibrillation, on the prediction of BP using rPPG is currently uncertain. We performed a study to better understand the relationship between rPPG and BP in a real-world sample of ambulatory patients from a cardiology clinic with established CVD or risk factors for CVD. We collected simultaneous rPPG, PPG, BP, ECG, and other vital signs data from 143 subjects while at rest, and used this data plus demographics to train a deep learning model to predict BP. We report that facial rPPG yields a signal that is comparable to finger PPG. Pulse wave analysis (PWA)-based BP estimates on this cohort performed comparably to studies on healthier subjects, and notably, the accuracy of BP prediction in subjects with atrial fibrillation was not inferior to subjects with normal sinus rhythm. In a binary classification task, the rPPG model identified subjects with systolic BP $\geq$ 130 mm Hg with a positive predictive value of 71\% (baseline prevalence 48.3\%), highlighting the potential of rPPG for hypertension monitoring. 


\end{abstract}


\maketitle  

\section{Introduction}

Hypertension remains the leading preventable cause of global morbidity and mortality from cardiovascular disease (CVD) including myocardial infarction, heart failure, and stroke. The number of adults age 30-79 with hypertension has approximately doubled between 1990 and 2019, from 600 million to over 1.2 billion. An estimated 1 in 3 adults have hypertension, of which only 1 in 5 have adequate BP control \cite{zhou2021worldwide}.

Improving hypertension management requires a multifaceted approach including development of novel technology to measure BP for the initial diagnosis and subsequent monitoring outside of brick-and-mortar healthcare settings. The gold standard method to measure BP is manual manometry with auscultation. In clinical practice, the most widely used method is automated oscillometric cuff pressure, in which a computer detects oscillations of blood flow at various cuff pressures to estimate systolic and diastolic BP and mean arterial pressure based on proprietary algorithms \cite{sharman2023automated}. Cuff-based BP methods have limitations such as high cost and limited availability of validated devices, poor patient adherence due to discomfort and inconvenience, and inaccurate measurements caused by improper cuff sizing and placement \cite{mukkamala2022cuffless, ishigami2023effects}. Consequently, there is a need for the development of new technologies to improve hypertension control.

Photoplethysmography (PPG) is an optical signal of a pulse waveform obtained by illuminating the skin with a light emitting diode (LED) and measuring the amount of reflected light using a photodiode, which detects blood volume changes in the peripheral microvasculature based on the light absorption properties of hemoglobin. PPG has been used widely in medicine for pulse oximetry, which measures pulse rate and blood oxygen saturation from a device typically attached to the fingertip or ear lobe. The consumer health industry has witnessed a renewed interest in PPG technology, primarily driven by the growing popularity of wearable devices. PPG-enabled algorithms and machine learning (ML) models have been studied for detection of atrial fibrillation \cite{perez2019large, lubitz2022detection}, estimation of blood glucose concentration and detection of diabetes based on heart rate variability \cite{zanelli2022diabetes}, vascular aging \cite{alastruey2023arterial}, and, with uncertainty, as a tool for BP monitoring \cite{mukkamala2023microsoft}. The clinical community is in the early stages of understanding how to utilize PPG-based wearable devices for patient care \cite{ginsburg2024key, spatz2024wearable}. Additionally, concerns remain regarding cost, accessibility, and adherence of wearable devices when used for tracking physical activity \cite{chan2022reporting}. 



Advances in camera technology have enabled a wide variety of health sensing, including body temperature, pulse rate, and respiratory rate~\cite{nowara2022seeing}. Whereas conventional PPG requires physical contact with skin, remote photoplethysmography (rPPG)~\cite{mcduff2023camera} utilizes a standard RGB camera as the photodetector for PPG measurement. Typically, ambient light is utilized and the camera is held from a distance to the subject.  rPPG offers potential for passive monitoring without the need to wear a sensor, and a lens that can capture more optical information from a larger skin surface area than a typical PPG photodiode. rPPG is considered highly scalable and promising as RGB cameras are commonly embedded in devices including smartphones and laptops. According to a recent report, the majority of the world's population owns a smartphone \cite{GSMA2023}.

The methods to measure BP using PPG can be generally categorized into two approaches:  1) Time-based methods include pulse transit time (PTT) and pulse arrival time (PAT). PTT ~\cite{ptt-1,ptt-2,ptt-3} is based on the time delay for a pressure wave to travel between proximal and distal arterial sites. The PTT approach has strong theoretical underpinnings based on the Bramwell-Hill equation, which relates PTT to pulse wave velocity and arterial compliance. The Wesseling model captures the relationship between arterial compliance and BP ~\cite{wesseling1993computation}. PAT measures the time delay between electrical ventricular contraction determined by an ECG and a distal arterial site, such as the wrist or finger. However, PAT is confounded by the variable delay between electrical and mechanical ventricular contraction, known as the pre-ejection period, which cannot be reliably measured without invasive means. PTT and PAT can change independently of BP due to variations in arterial compliance related to smooth muscle tone and age-related arteriosclerosis, thus requiring periodic calibration ~\cite{mukkamala2022cuffless}. 2) Pulse wave analysis (PWA) is a method used to estimate BP by extracting features from a pulse waveform. Unlike PTT and PAT, PWA has weaker theoretical underpinnings as the small arteries interrogated by PPG are viscoelastic \cite{ptt-1}. PWA is attractive because it offers a single-device solution, whereas PTT and PAT typically requires multisite and multimodal measurements. While researchers previously relied on simple regression and conventional ML of handcrafted waveform features for PWA, advances in deep learning have enabled greater waveform feature comparisons with relative ease \cite{van2023improving}. 

To date, studies of rPPG-based BP estimation have shown mixed results \cite{curran2023camera, NATARAJAN2022117, steinman2021smartphones, molinaro2022contactless, selvaraju2022continuous}. Significant advancements in signal processing, feature extraction, and ML have been achieved by researchers primarily from the biomedical, computer science, and electrical engineering disciplines. However, previous studies have generally lacked clinical perspectives and have been subject to several major limitations which this study attempts to address. With few exceptions \cite{schrumpf2021assessment, heiden2022measurement, van2023improving,wu2022facial}, study populations have been limited to normotensive subjects. Testing models exclusively on subjects with a limited range of BP raises concern for the validity of BP prediction, which could simply be a prediction of the mean value  \cite{mukkamala2022cuffless}. Additionally, participants are generally young and healthy without CVD or with few cardiometabolic risk factors, which considerably limits the generalizability of findings. For example, arterial stiffness is known to affect pulse wave velocity and waveform shape, crucial variables for PTT and PWA-based methods of BP prediction \cite{chiarelli2019data}. Factors that can affect arterial stiffness such as aging must be considered. Similarly, medications routinely prescribed for management of CVD have a variety of effects on the cardiovascular system, such as reduced cardiac contractility and decreased systemic vascular resistance, and therefore may affect PPG-based BP prediction. Irregular rhythms such as atrial fibrillation are common in the general population, yet the impact of heart rhythm on BP prediction has not been thoroughly investigated. In the two largest rPPG studies of over 1000 subjects, heart rhythm was not recorded with simultaneous ECG acquisition~\cite{luo2019smartphone, van2023improving}; another rPPG study included a minority of subjects with a history of atrial fibrillation, but a subgroup analysis was not done~\cite{wu2022facial}; in a study of a PPG device available to consumers, subjects with arrhythmia were specifically excluded \cite{Aktiia_tan2023evaluation}. Abnormal heart rhythms may impact diastolic filling time, subsequently affecting stroke volume, pulse pressure, and potentially altering the peripheral pulse waveform used in predicting BP prediction through PWA, or the time delay in PTT calculation. Without simultaneous ECG, the heart rhythm cannot be determined and thus a comparison of BP prediction by rhythm type has not been previously studied. 

Although other studies have investigated the performance of different architectures that underlie the predictive ML models, the focus of this study was to provide clinical context, which relates to the features used to train the model. We compared the performance of a single deep learning model architecture (a 1-dimensional convolution neural network plus transformer) trained on different sets of features (``baseline'' composed of demographics, medications, and medical history, ``rPPG" or "PPG" composed of pre- and post-processed pulse waveforms, and a ``hybrid'' combination of rPPG/PPG and baseline features) to determine the information gain from rPPG features compared to baseline features alone. Although we are not attempting to validate a device in this study, within the framework for cuffless BP devices described by the European Society of Hypertension (ESH)\cite{stergiou2023european}, the rPPG and PPG models are "calibration-free" in that the oscillometric cuff BP was used as ground truth data for model training only -- subsequent individual cuff calibration was not part of this study design; the hybrid model is a "demographic-calibrated" model based on the individual's demographic (and medical history) features. 

In summary, the objectives of this study were to: 1) Create a multi-source biosignal dataset (N = 143) of rPPG, PPG, BP, ECG, RR, and SpO2 data in a static setting from a real-world ambulatory population with CVD or CV risk factors; 2) Use PWA of rPPG and PPG, medical history, and demographic features to train and compare deep learning models to predict BP; 3) To test if models trained with subjects in active arrhythmia (atrial fibrillation, frequent ectopy, or paced rhythm) have higher BP prediction error than models trained with subjects in normal sinus rhythm; and 4) Test a use case for rPPG to distinguish between subjects with systolic BP  $\geq$ or $<$ 130 mm Hg) from a model trained on a binary classification task.




\section{Results}

\subsection{Subject Demographics and Characteristics}

\begin{table}[h]
    \centering
\setlength{\tabcolsep}{6pt}
\begin{center}
\adjustbox{max width=\textwidth}{
\begin{tabular}{lclclc}
\bottomrule[1.5pt]
Characteristic & n (\%)/mean ($\mp$SD) & Medical History & n (\%) & Medications & n (\%) \\ 
\midrule
\grayrow
Total patient visits & N = 143 & Dyslipidemia & 110 (76.9\%) & Antihypertensives & 107 (74.8\%) \\[1.2pt]
\quad Female & 55 (39\%) & Hypertension & 101 (71\%) & \quad ACEi/ARB & 72 (50\%) \\[1.2pt]
\grayrow
\quad Male & 88 (62\%) & Diabetes mellitus type II & 54 (38\%) & \quad Beta blocker & 73 (51\%) \\[1.2pt]
Race/Ethnicity* & & Coronary artery disease & 52 (36\%) & \quad Ca channel blocker & 42 (29\%) \\[1.2pt]
\grayrow
\quad White/Non-Hispanic/Latino & 114 (80\%) & Atrial fibrillation or flutter & 39 (27\%) & \quad Thiazide & 27 (19\%) \\[1.2pt] 
\quad White/Hispanic/Latino & 1 (1\%) & Sleep apnea & 30 (21\%) & \quad MRA & 10 (7\%) \\[1.2pt] 
\grayrow
\quad Asian & 21 (15\%) & Valvular heart disease & 22 (15\%) & Antiplatelet/Anticoag. &  \\[1.2pt] 
\quad Black or African-American & 1 (1\%) & Cardiomyopathy & 16 (11\%) & \quad Asiprin & 56 (39\%) \\[1.2pt] 
\grayrow
\quad Unknown or unanswered & 6 (4\%) & Congestive heart failure & 14 (10\%) & \quad Warfarin & 39 (27\%) \\[1.2pt] 
Age (years) & 69 ($\mp$11.7) & Chronic kidney disease & 13 (9\%) &  Statin & 98 (69\%) \\[1.2pt] 
\grayrow
BMI (kg/m$^2$) & 28.4 ($\mp$5.6) & Pacemaker & 12 (8\%) &  Diuretic & 9 (6\%) \\[1.2pt] 
Systolic BP (mmHg) & & & & Antiarrhythmic & 6 (4\%) \\[1.2pt] 
\grayrow
\quad Time 0 & 134.1 ($\mp$19.5) & & & PDE$_{5}$i & 5 (4\%) \\[1.2pt] 
\quad Time $\sim$3 minutes & 130.4 ($\mp$17.9) & & & SGLT2i & 3 (2\%)  \\[1.2pt] 
\grayrow
\quad Time $\sim$6 minutes & 129.9 ($\mp$18.1) & & &  ARNi & 3 (2\%)  \\[1.2pt] 
Diastolic BP (mmHg) & & & & \\[1.2pt] 
\grayrow
\quad Time 0 & 79.5 ($\mp$9.1) & & & & \\[1.2pt] 
\quad Time $\sim$3 minutes & 77.9 ($\mp$9.4) & & & & \\[1.2pt] 
\grayrow
\quad Time $\sim$6 minutes & 77.6 ($\mp$8.4) & & & & \\[1.2pt] 
\bottomrule[1.5pt]
\end{tabular}}
\end{center}
    \caption{\textbf{Subject Characteristics.} Summary of subject demographics, relevant medical history, and medications. ACEi = angiotensin-converting enzyme inhibitor, ARB = angiotensin II receptor blocker, ARNi = angiotensin receptor/neprilysin inhibitor, BMI = body mass index, BP = blood pressure, MRA = mineralocorticoid receptor antagonist, PDE$_{5}$i = phosphodiesterase-5 inhibitor, SGLT2i = sodium-glucose cotransporter-2 inhibitor}
    \label{tab:demographics}
\end{table}


\subsection{Model Comparisons}
\begin{figure}
    \caption{\textbf{rPPG Blood Pressure Estimation Pipeline.} The rPPG signal is preprocessed frame-by-frame from the video. After preprocessing, five consecutive beats with good quality are selected by template matching. The selected beat sequence is encoded with a conformer and features embeddings are appended with embeddings from demographic and clinical history features. Finally, a multi-layer perceptron outputs the blood pressure estimate. Blood pressure measurements from the oscillometric cuff are used as ground truth data for model training, not for individual calibration. 
    MLP = multi-layer perceptron, BMI = body mass index.}
    \includegraphics[width=\linewidth]{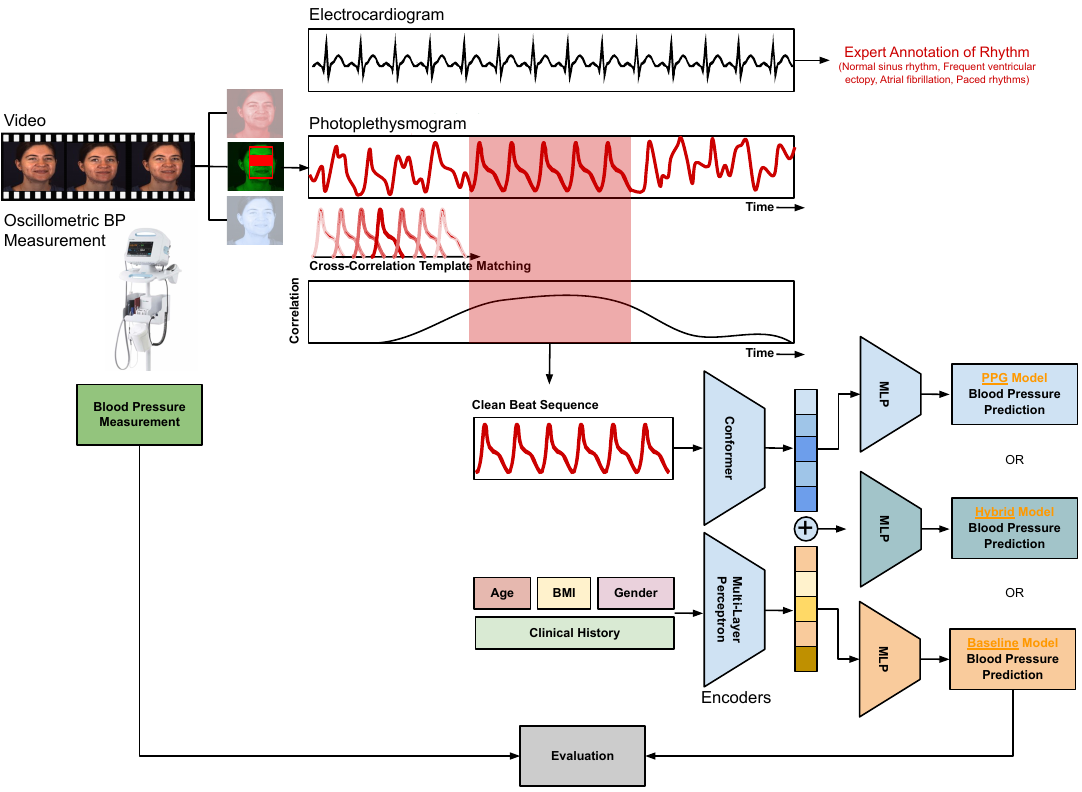}
    \label{fig:architecture}
\end{figure}

A baseline model was developed to serve as a benchmark for comparison, trained on subject demographics, past medical history, and medications (Table~\ref{tab:demographics} and Figure~\ref{fig:architecture}). The model processed a total of 37 features through a multi-layer perceptron (MLP) neural network, yielding an output of 64 features. Next, models trained exclusively on PPG or rPPG signals, named "PPG models," analyzed sequences of pulse segments, each padded to 512 elements, and employed an MLP to produce a latent vector of 128 dimensions. Further advancing the analytical framework, "Hybrid models" integrated the insights from the baseline model with those from the PPG models. As illustrated in Figure~\ref{fig:architecture}, this synergistic approach generated a latent vector of 192 dimensions, encapsulating a comprehensive representation of the physiological and demographic factors pertinent to BP estimation.

For each model, the Pearson correlation, mean absolute error of the residuals (MAE), and standard deviation (SD) of predicted systolic blood pressure (SBP) compared to the ground truth cuff SBP are presented in the upper section of Table~\ref{tab:sys_bp_results_table}. Due to the non-normal distribution of MAE values, a Mann-Whitney U test was applied for comparing models (\textit{p} $<$ 0.01).




There was a consistent increase in Pearson correlation and reduction in MAE between the baseline, rPPG or PPG, and respective Hybrid models. For pairwise comparisons of PPG to baseline and Hybrid to baseline (marked by $^*$ in Table~\ref{tab:sys_bp_results_table}), the respective PPG and Hybrid models exhibited marginally lower mean absolute errors (MAE) for SBP prediction (rPPG vs. baseline MAE: $-$3.35 mm Hg, rPPG Hybrid vs. baseline MAE: $-$3.61 mm Hg, PPG vs. baseline MAE: $-$2.81 mm Hg, PPG Hybrid vs. baseline MAE: $-$3.12 mm Hg; \textit{p} $<$ 0.01). However, for pairwise comparisons of Hybrid to PPG, in which baseline features were integrated into the PPG models, there was no significant difference in MAE for either rPPG or PPG. For pairwise comparisons of rPPG and PPG models (marked by $^\lambda$ in Table~\ref{tab:sys_bp_results_table}), a statistically significant (but clinically insignificant) difference was observed in the MAE for both PPG and Hybrid models, in favor of rPPG (rPPG vs. PPG MAE: $-$0.54 mm Hg, rPPG Hybrid vs. PPG Hybrid MAE: $-$0.49 mm Hg; \textit{p} $<$ 0.01).





\begin{table}[h!]
    \centering
\setlength{\tabcolsep}{6pt}
\adjustbox{max width=\textwidth}{
\begin{tabular}{rcccccccc}
\bottomrule[1.5pt]
\multirow{2}{*}{Model/Subgroup} & \multicolumn{3}{c}{Face rPPG} & \multicolumn{3}{c}{Finger PPG} & \multicolumn{2}{c}{Systolic BP Cuff} \\ 
 & \textit{r}  & MAE (mm Hg)  & SD & \textit{r}  & MAE (mm Hg) & SD & Mean (mm Hg)  & SD  \\ 
\midrule
All Sessions  \\[1.2pt]
\quad Baseline Model (n$_{rPPG}$ = 243, n$_{PPG}$ = 245) &  0.261 & 14.96 & 18.57 & 0.261 & 14.96 & 18.57 \\[1.2pt]
\quad PPG Model (n$_{rPPG}$ = 243, n$_{PPG}$ = 245) & 0.499	& 11.61$^{*,\lambda}$ & 14.63 & 0.475 & 12.15$^{*,\lambda}$ & 15.17 & 132.67 & 17.03 \\[1.2pt]
\quad Hybrid Model (n$_{rPPG}$ = 243, n$_{PPG}$ = 245) & \textcolor{black}{\textbf{0.541}} & \textcolor{black}{\textbf{11.35$^{*,\lambda}$}} & \textcolor{black}{\textbf{14.20}} & \textcolor{black}{\textbf{0.501}} & \textcolor{black}{\textbf{11.84$^{*,\lambda}$}} & \textcolor{black}{\textbf{15.18}}  \\[1.2pt]
\midrule
Rhythm Groups  \\[1.2pt]
\quad NSR (n$_{rPPG}$ = 186, n$_{PPG}$ = 186) & 0.457 & 12.29 & 15.30 & 0.382 & 13.17 & 16.24 & 131.75 & 17.68 \\[1.2pt]
\quad AF (n$_{rPPG}$ = 27, n$_{PPG}$ = 28) & 0.550 & 10.16$^{\dagger}$ & 12.48 & 0.569 & 9.30$^{\dagger}$ & 12.10 & 125.10 & 14.27 \\[1.2pt]
\quad Freq. Ectopy (n$_{rPPG}$ = 8, n$_{PPG}$ = 7)& 0.503 & 14.21 & 18.13 & 0.447 & 15.21 & 18.41 & 137.35 & 19.00  \\[1.2pt]
\quad Paced (n$_{rPPG}$ = 15, n$_{PPG}$ = 17)& 0.210 & 13.33$^{\dagger}$ & 13.26 & 0.705	& 9.48$^{\dagger}$ & 9.52 & 137.84 & 12.92  \\[1.2pt]
\bottomrule[1.5pt]
\end{tabular}}
\footnotesize
$^{*}$ = sig. difference (p $<$ 0.01) compared to Baseline Model, 
$^{\dagger}$ = sig. difference (p $<$ 0.01) compared to NSR,
$^{\lambda}$ = sig. difference (p $<$ 0.01) between respective rPPG and PPG models
    \caption{\textbf{Systolic Blood Pressure Estimation Results.} Pearson correlation (\textit{r}), mean absolute error (MAE) and standard deviation (SD) in systolic blood pressure estimation from facial based remote PPG (Face rPPG) and finger based contact PPG (Finger PPG). The respective number of sessions for rPPG and PPG groups are shown.
    BP = blood pressure, NSR = normal sinus rhythm, AF = atrial fibrillation}
    \label{tab:sys_bp_results_table}
\end{table}

\subsection{Comparison of BP by Heart Rhythm}

\begin{figure}
\caption{\textbf{Comparative Segments of PPG, Remote PPG, and ECG Across Different Heart Rhythms.}}
\centering
\includegraphics[width=0.9\linewidth]{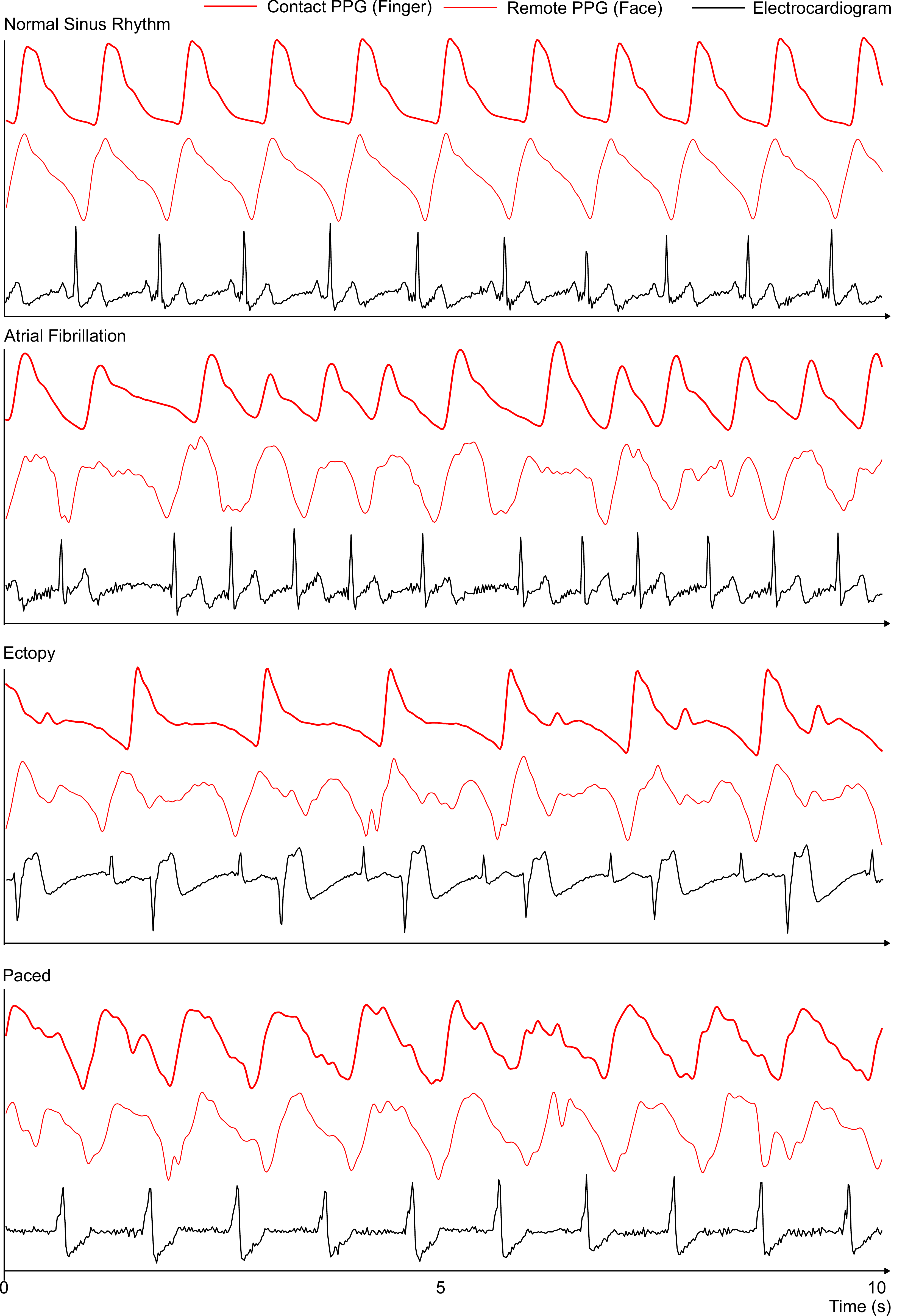}
\label{fig:ppg_segment}
\end{figure}

\begin{figure}
    \caption{\textbf{Correlation and Bland-Altman Plots.} Systolic blood pressure estimation results from face rPPG.}
    \subfloat[]{\includegraphics[height=0.4\linewidth]{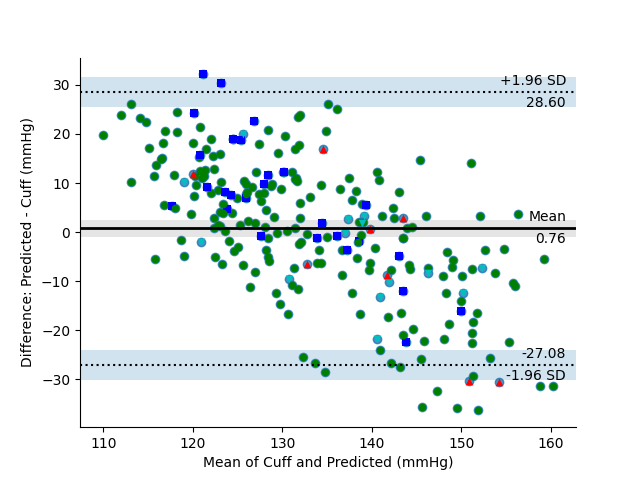}}
    \subfloat[]{\includegraphics[height=0.4\linewidth]{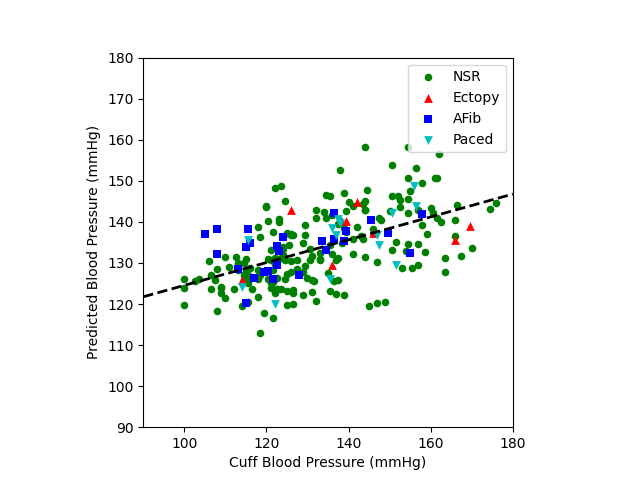}}
    \label{fig:blandaltman}
\end{figure}

SBP prediction results by heart rhythm group are detailed in the lower section of Table \ref{tab:sys_bp_results_table}. Figure \ref{fig:ppg_segment} shows qualitative comparisons of PPG, rPPG, and ECG segments from each rhythm group. Figure \ref{fig:blandaltman} presents Bland-Altman and scatter plots for SBP prediction by rhythm group. Only results from the PPG models are reported for each heart rhythm group, because as described above the Hybrid models did not demonstrate significantly different SBP prediction errors compared to PPG models when trained across all subjects. In addition, an analysis of the significance of clean versus noisy training revealed no significant differences in SBP prediction accuracy. Therefore, only the results from training with all subjects (i.e., noisy training) are reported. Due to the non-normal distribution of MAE, a Mann-Whitney U test was utilized for pairwise group comparisons (\textit{p} < 0.01). For both rPPG and PPG, the atrial fibrillation (AF) group exhibited lower SBP prediction errors compared to the normal sinus rhythm (NSR) group (rPPG MAE: 10.16 vs. 12.29 mm Hg, PPG MAE: 9.30 vs. 13.17 mm Hg; \textit{p} $<$ 0.01). The frequent ectopy groups had a trend toward increased error compared to NSR groups that did not meet statistical significance (rPPG MAE: 14.21 vs. 12.29 mm Hg, \textit{p} = 0.12; PPG MAE: 15.21 vs. 13.17 mm Hg, \textit{p} = 0.31). The paced rhythm group showed increased error compared to the NSR group for rPPG (MAE: 13.33 vs. 12.29 mm Hg, \textit{p} $<$ 0.01), but decreased error for PPG (MAE: 9.47 vs. 13.17 mm Hg, \textit{p} = 0.35).

\subsection{Binary Classification of Systolic Blood Pressure: A Model for Hypertension Screening}

A binary classification task was conducted using a model pipeline similar to the one depicted in Figure~\ref{fig:architecture}, with the key difference being the output categorized as either SBP $\geq$ or $<$ 130 mm Hg. The results for the baseline, rPPG, and rPPG Hybrid models are illustrated in Figure~\ref{fig:binary} and in Supplemental Table~\ref{tab:my_label}. The prevalence (or pre-test probability) of SBP $\geq$ 130 mm Hg in this cohort was 48.3\%. The baseline model showed near random BP prediction (accuracy = 51.4\%). Both the rPPG and Hybrid models surpassed the baseline model in performance, exhibiting similar accuracy (accuracy = 63.8\% and 67.5\%, respectively; McNemar's test, \textit{p} $<$ 0.05). The rPPG and Hybrid models also demonstrated comparable positive predictive values (70.8\% and 70.9\%, respectively). The rPPG model had only moderate sensitivity (50.4\%), which improved with the integration of baseline features into the Hybrid model (sensitivity = 62.4\%).

\section{Discussion}

\subsection{Model Comparisons}

\begin{figure}
\centering
    \caption{\textbf{Binary Classification of SBP.} Classifying hypertensive vs non-hypertensive subjects defined as SBP $\geq$ or $<$ 130 mm Hg, respectively (based on the 2017 ACC/AHA BP guidelines for the general population.) The prevalence (or pre-test probability) of SBP $\geq$ 130 mm Hg in this cohort was 48.3\%. The hybrid model produces the most accurate model with the highest sensitivity.}
    \includegraphics[width=0.5\linewidth]{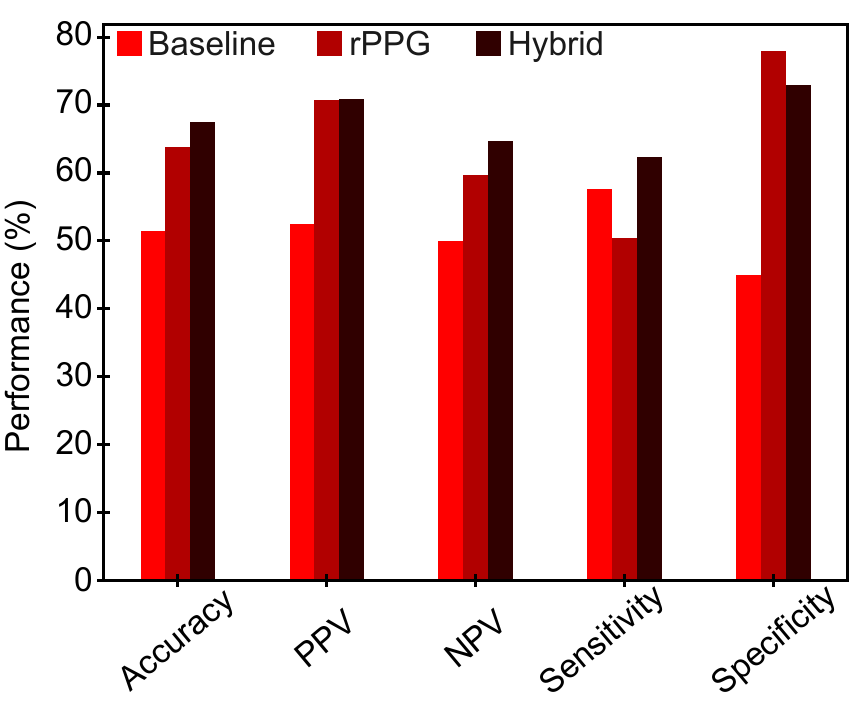}
    \label{fig:binary}
\end{figure}

In general, combining baseline features with the PPG or rPPG signals to create a Hybrid model only modestly improved SBP prediction compared to models relying solely on PPG or rPPG features. Our findings suggest PPG and rPPG contain valuable hemodynamic information that facilitates SBP estimation beyond what can be inferred from demographic data and medical history alone. The extent to which a real-world clinical model incorporates baseline features should be carefully considered and tailored to the intended patient population. 

The baseline model sets a benchmark for comparison, which is essential when making claims of BP prediction from ML models combining physiological measurements like rPPG and demographic data, the latter of which has been shown to correlate with BP \cite{mukkamala2022cuffless}. In comparison to other studies, Van Putten et al. developed a baseline model that included age, sex, height, and weight \cite{van2023improving}.  Heiden et al. did not specify a separate baseline model, instead reporting a hybrid-type model that integrated age, sex, Fitzpatrick skin type, and rPPG features \cite{heiden2022measurement}. Luo et al. expanded their baseline model to include age, sex, height, weight, skin tone, race, and heart rate \cite{luo2019smartphone}. In our study, we chose to omit heart rate from the baseline model for two main reasons: first, the association between heart rate and BP is context-dependent; and second, we aimed to distinguish physiological from non-physiological features. Our baseline and hybrid models incorporate demographic data and distinctive features of CVD and current medications. Knowledge of the diagnosis of hypertension, or the use of vasoactive medications that could influence PPG waveform dynamics, might enhance model accuracy. However, despite inclusion of these features, our baseline model showed surprisingly poor accuracy in SBP prediction. This would explain why the Hybrid models generally did not have better SBP prediction accuracy than PPG models.  

CVD encompasses a variety of pathologies, each potentially interacting with PPG morphology in distinct ways. 
A recent study reported a Bayesian ML model using only time-domain features of PPG that was able to classify healthy subjects from subjects with CVD with moderate accuracy (66.12\%), and also distinguished different types of CVD including acute coronary syndrome, stroke, heart failure, atrial fibrillation, and deep vein thrombosis (accuracy = 62.28\%)~\cite{al2023identification}. This suggests there may be unique differences in PPG morphology associated with specific CVD types. It is unclear to what extent these differences impact PPG-based BP prediction. Wu et. al reported a higher MAE in rPPG-based BP prediction in a subgroup analysis of subjects with CVD; however, details of the CVD group are limited, and it is uncertain if these findings are due to differences in BP distribution and testing environments rather than CVD specifically~\cite{wu2022facial}. Interestingly, our CVD cohort demonstrated a similar correlation between rPPG and BP (\textit{r} = 0.541) compared to healthier cohorts, based on a review which calculated the unweighted average of the highest correlation coefficient for BP prediction across studies (\textit{r} = 0.54)  \cite{NATARAJAN2022117}. This suggests that CVD, as represented in this patient population, does not substantially affect rPPG-based BP prediction in comparison to healthier cohorts. Nevertheless, the broad error margin in BP prediction constrains the interpretation of these results, and the absence of a control group in our study without CVD or CV risk factors limits a definitive comparison. 




Previous studies have explored a range of model architectures for BP estimation, including multiple linear regression models utilizing handcrafted rPPG waveform features \cite{Microsoft_Aurora_Project_mieloszyk2022}, ensemble ML algorithms such as random forests \cite{heiden2022measurement}, and stacked ensembles \cite{van2023improving}. Although the focus of our research was not model architecture design, the progression towards our final model merits attention. Initially, we examined multiple linear regression using known waveform features, which yielded unsatisfactory predictive accuracy. Through subsequent iterations, we transitioned to an ensemble approach combining CNN, vision transformer, and MLP architectures. 


\subsection{Remote vs Contact PPG}


Prior research indicates that the waveform quality from rPPG is typically not as robust as contact PPG, which can make the prediction of heart rate and BP more challenging \cite{mukkamala2022cuffless}. In contrast, our findings indicate that rPPG is non-inferior to PPG for BP prediction, however the large error margins may limit the validity of these results. Of note, the non-inferior performance of rPPG compared to PPG was observed despite the apparently higher noise of rPPG seen by qualitative assessment (Figure \ref{fig:ppg_segment}). We confirmed this observation by performing a quantitative assessment of our PPG signal quality compared to a dataset provided by Kachuee et al. from the UCI Machine Learning Repository \cite{kachuee2015cuff}, as detailed in the Supplemental Results.




\subsection{Comparison of BP Prediction by Heart Rhythm}


One of our objectives was to assess if the presence of atrial or ventricular ectopy, atrial fibrillation or flutter, or paced rhythm influences BP prediction accuracy compared to normal sinus rhythm. We hypothesized that premature contractions, atrial fibrillation, or paced rhythm, by affecting the timing of atrial contraction or ventricular diastolic filling time, would cause variations in stroke volume and blood volume pulse amplitude, leading to increased error in BP prediction compared to normal sinus rhythm. Contrary to our expectations, the results were mixed. In particular, the atrial fibrillation group demonstrated lower error in SBP prediction than the normal sinus rhythm group. Without a specific physiologic mechanism to support this finding, there may be limitations in methodology. The mean and standard deviation of ground truth SBP differed across rhythm groups, which likely contributes to the observed variations in prediction error. Given the relatively small sample sizes of groups with arrhythmia, these findings should be interpreted with caution. The influence of template matching on PPG signal selection, which may differ between rhythm groups, could also affect the prediction error. 



\subsection{Binary Classification of Systolic Blood Pressure}

There is reasonable concern that current PPG-based BP measurement devices (including those based on PWA) are not as reliable as an automated oscillometric device in real-world conditions~\cite{mukkamala2023microsoft}, which may imply similar concerns for rPPG. Instead of performing intermittent or continuous BP measurement, a more pragmatic use for PPG and rPPG could be to screen for hypertension or detect BP above a threshold following diagnosis. By classifying individuals into binary categories (hypertensive or non-hypertensive), rPPG could prompt users to verify their BP using validated devices, such as a cuff-based BP monitor at home. This approach is similar to the PPG-based wearable devices that screen for atrial fibrillation and alert users to seek medical attention for diagnostic confirmation via ECG or cardiac ambulatory monitor~\cite{perez2019large,lubitz2022detection}. 



Our results show promise that rPPG can be used to screen for hypertension or subsequently for monitoring following diagnosis. We tested several models on a binary classification task to distinguish between SBP classes. Although BP targets are not uniform across clinical guidelines \cite{BPtargetcon}, we used a cut-off of SBP $\geq$ 130 mm Hg based on the 2017 ACC/AHA BP Guideline for diagnosis of "stage 1 hypertension" in the general population \cite{whelton20182017} and the 2023 ESH BP Guideline definition for "high-normal" BP which is known to correlate with adverse CV morbidity \cite{mancia20232023}. First, we determined the pre-test probability: 48.3\% of total sessions (two per study participant) had an average SBP $\geq$ 130 mm Hg. Next, we tested a baseline model for control using only demographic features, which showed ambivalent SBP classification (accuracy = 51.4\%). We then tested the rPPG and Hybrid models, which showed accuracy of 63.8\% and 67.5\%, respectively, and positive predictive value of 70.8\% and 70.9\%, respectively. The sensitivity of the rPPG model was disappointingly low (50.4\%), but improved with the integration of baseline features into the Hybrid model (sensitivity = 62.4\%). For context, in a related study by Van Putten et al., their model had a sensitivity of 79\% for detecting hypertension, likely in part due to the higher cut-off for SBP $>$ 140 mm Hg \cite{van2023improving}. It is an interesting observation that demographic features had a greater effect on improving sensitivity for classifying higher SBP than improving accuracy of exact SBP measurement.


\section{Limitations}

This study was exploratory in design and was not intended to validate rPPG or a specific device for BP measurement. Cuffless BP measurement includes many novel technologies, of which rPPG is one of the least studied. A major limitation to all cuffless BP methods and devices has been a lack of a standard protocol for validation. In 2018, the Advancement of Medical Instrumentation/European Society of Hypertension/International Organization for Standardization (AAMI/ESH/ISO) published a protocol that was namely intended for cuff-based devices \cite{stergiou2018universal}. The ESH has previously advised against the use of cuffless BP devices in clinical practice, based largely on the lack of standard protocols for validation, especially for devices that require periodic cuff-calibration \cite{stergiou2022european}. Post-market studies of regulatory-cleared devices and applications using PPG that require periodic cuff calibration have performed poorly when compared to 24-hour ambulatory BP monitoring or tracking BP over days to weeks due to drift away from reference standard \cite{Microsoft_Aurora_Project_mieloszyk2022, Aktiia_tan2023evaluation, holyoke2021web}. Therefore, the ESH recently released updated recommendations for the validation of cuffless BP measuring devices (the "ESH standard") \cite{stergiou2023european}. Even so, several mobile apps are on the market that claim to measure BP using rPPG from a smartphone camera, but have not been validated using the ESH standard~\cite{attune,binah,nervotec,rajan2023accuracy}. 

Within the framework published in the ESH standard, this study design would be considered cuff calibration-free, and therefore the prior AAMI/ISO/ESH standard may apply. However, there still remain several limitations. Foremost, other than a select few subjects (see Supplemental Results), we did not sufficiently test the accuracy of the model for BP prediction over time. We also do not know the accuracy for BP prediction over time for a cuff calibration-free rPPG model compared to a cuff-calibration rPPG model. Future studies should address these questions. We used automated oscillometric cuff BP as the reference standard, not manual auscultatory BP as recommended. Of the several validation tests set forth in the prior AAMI/ISO/ESH and newer ESH standards, this study only addresses the static test; future studies should also address positional (orthostatic) testing, the effect of antihypertensive treatment, the effect of sleep compared to awake states, the effect of exercise, and the sustained accuracy just prior to cuff-calibration if used. 

Previous studies have raised concern for lower accuracy of pulse oximetry using contact PPG or rPPG in individuals with darker skin tones \cite{sjoding2020racial, gudelunas2022low, boonya2021monte, nowara2020meta}. In this study, participants were recruited using convenience sampling, without predefined criteria for demographics, medical history, heart rhythm, or BP. We did not classify subjects by Fitzpatrick skin type, due to the knowledge that the vast majority of subjects self-identified as Non-Hispanic White or Non-Hispanic Asian, limiting the range of skin tones in model development. Notably, in the study by Heiden et al., inclusion of Fitzpatrick skin type did not impact BP prediction from an rPPG-based model, despite a diverse range of skin types studied \cite{heiden2022measurement}. However, further investigation into the impact of skin tone on rPPG-based BP estimation specifically is needed.

Of 134 unique subjects, there were 9 return subjects. All data (that was not excluded per the Methods) was used for model training; therefore, there is a slight bias in the model for these subjects. 

A smartphone camera offers great potential for BP monitoring with rPPG. We used a DSLR camera at 60 fps to acquire more image data than likely needed. A  smartphone camera should be used in future studies to determine generalizability of results. 

\section{Conclusions}

Our study adds to a growing body of literature describing PPG- and rPPG-based methods to estimate BP. We demonstrate that rPPG-based BP estimation in a cohort of patients with established CVD or risk factors for CVD was comparable to studies in healthier cohorts. Surprisingly, we found that BP prediction using facial rPPG was slightly more accurate than finger PPG. BP prediction using rPPG was robust even in the presence of arrhythmia such as atrial fibrillation and frequent atrial or ventricular ectopy. We also explored a pragmatic approach to hypertension monitoring in an ambulatory population, where rPPG was used to classify an individual as hypertensive or not (defined as systolic BP $\geq$ or < 130 mm Hg, respectively), rather than perform exact BP measurement. Next steps should include a real-world study whereby subjects are intermittently monitored using rPPG from a smartphone camera and are alerted to measure BP using a cuff-based device at home. Future studies should expand upon the range of BP,  study environments, and diverse skin tone representation. Validation of this technology will also require adherence to standard protocols such as the ESH statement on cuffless BP devices \cite{stergiou2023european}.

\section*{Data Availability}

The data for this study included video recordings of subjects' faces, personal health records, and vital signs. Due to the particularly sensitive nature of facial video recording, per our IRB and subject consent the data is not available to researchers other than the listed authors of this study. 


\section*{Code Availability}

Code for our analysis is available on GitHub:
https://github.com/ubicomplab/rPPG-BP





\balance
\clearpage



\appendix
\section{Supplemental Results}

\subsection{Binary Classification Task to Screen for Hypertension}

\begin{table}[h!]
    \centering
\begin{tabular}{cccc}
\bottomrule[1.5pt] 
Classification & Baseline Model & PPG Model & Hybrid Model \\ 
\midrule
Accuracy & 51.4\% & 63.8\% & 67.5\% \\
PPV & 52.6\% &	70.8\% & 70.9\% \\
NPV & 50.0\% & 59.7\% & 64.7\% \\
Sensitivity & 57.6\% & 50.4\% & 62.4\% \\
Specificity & 44.9\% & 78.0\%  & 72.9\% \\
\bottomrule[1.5pt]
\end{tabular}
    \caption{\textbf{Hypertension Screening.} Results of a model classifying subjects as hypertensive (SBP $\geq$ 130 mmHg) or not (SBP $<$ 130 mm Hg). The prevalence (or pre-test probability) of SBP $\geq$ 130 mm Hg in this cohort was 48.3\%.}
    \label{tab:my_label}
\end{table}

\subsection{Exploratory Analysis: BP Prediction over Time}


The ESH recently emphasized the need to validate the ability of cuffless BP devices to track BP over time \cite{stergiou2023european}. Two prominent studies of approved PPG-based BP devices both demonstrated poor performance of BP prediction over days to weeks~\cite{mukkamala2023microsoft, Aktiia_tan2023evaluation}. Notably, both studies were based on contact PPG, not rPPG. 

Although not part of our initial study design, 9 patients returned for 3-month follow up and enrolled as repeat subjects. For exploratory analysis, we separately trained and tested a Hybrid model from these 9 subjects and found modest performance for BP prediction over 3 months (MAE 7.7 +/- SD 8.7 mm Hg, r = 0.675). Although limited by small sample size, these findings encourage further investigation.

\subsection{PPG Signal Quality}

We evaluated the quality of finger PPG signals in our dataset by comparing our signals to the Cuff-Less Blood Pressure Estimation dataset \cite{kachuee2015cuff}, sourced from the University of California Irvine Machine Learning Repository (UCI dataset). This comparison was based on skewness, which previous research by Elgendi \cite{elgendi2016optimal} identified as an optimal metric for assessing PPG signal quality. The skewness of our finger PPG signals was 0.51, closely matching the 0.56 reported for the UCI dataset. In addition to skewness, kurtosis and signal-to-noise ratio (SNR) were also utilized to evaluate PPG signal quality. Our finger PPG exhibited a kurtosis of -0.95, compared to -1.02 for the UCI dataset. According to these metrics, the quality of our finger PPG signals is on par with the UCI dataset, which is encouraging for our findings that rPPG performed comparably (and slightly better) than PPG for BP prediction.

\begin{figure}
    \caption{\textbf{Box Plots.} A comparison of the PPG signal quality from our dataset to the UCI dataset from Kachuee et al. ~\cite{kachuee2015cuff}.}
    \subfloat[]{\includegraphics[height=0.4\linewidth]{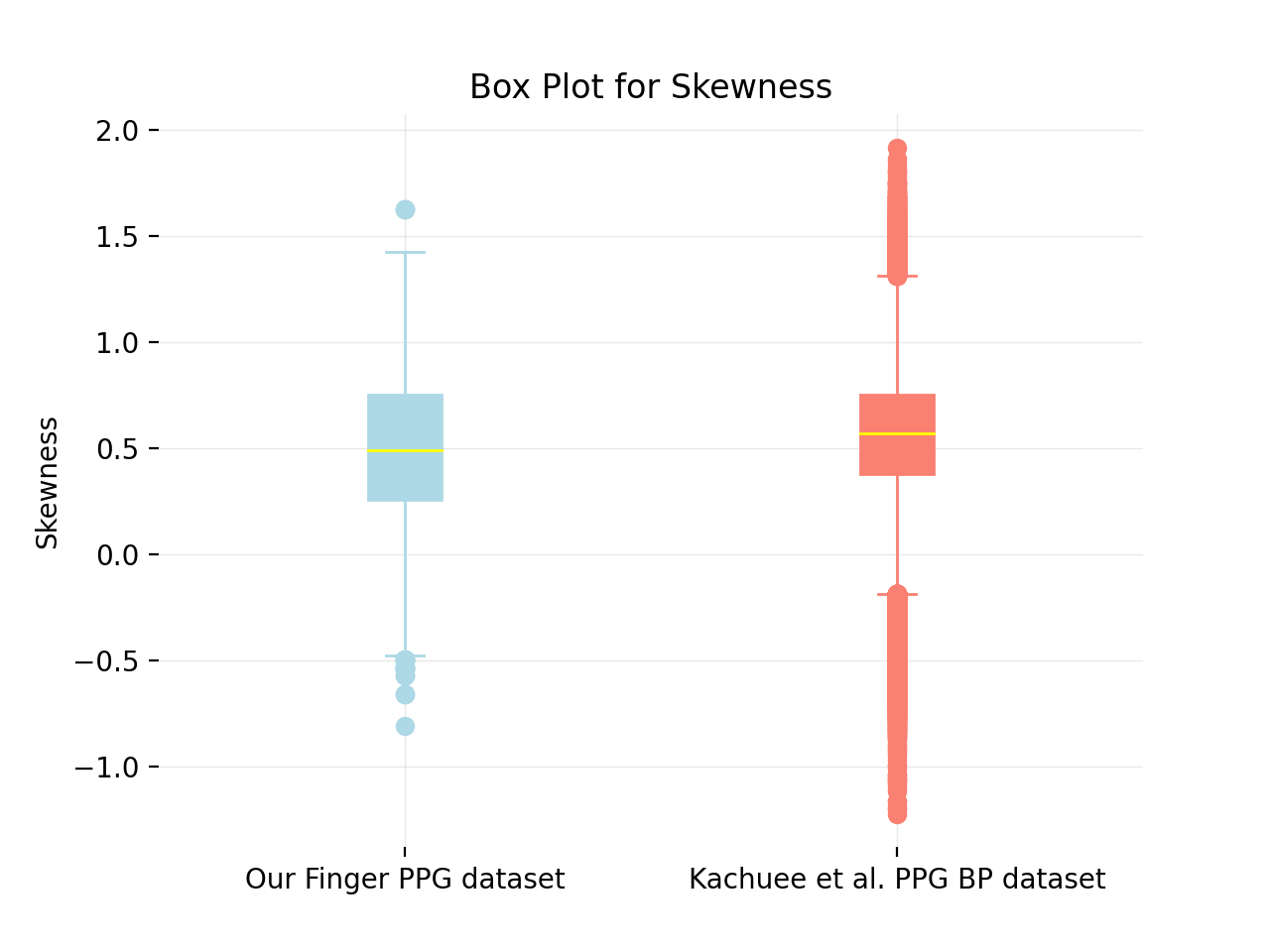}}
    \subfloat[]{\includegraphics[height=0.4\linewidth]{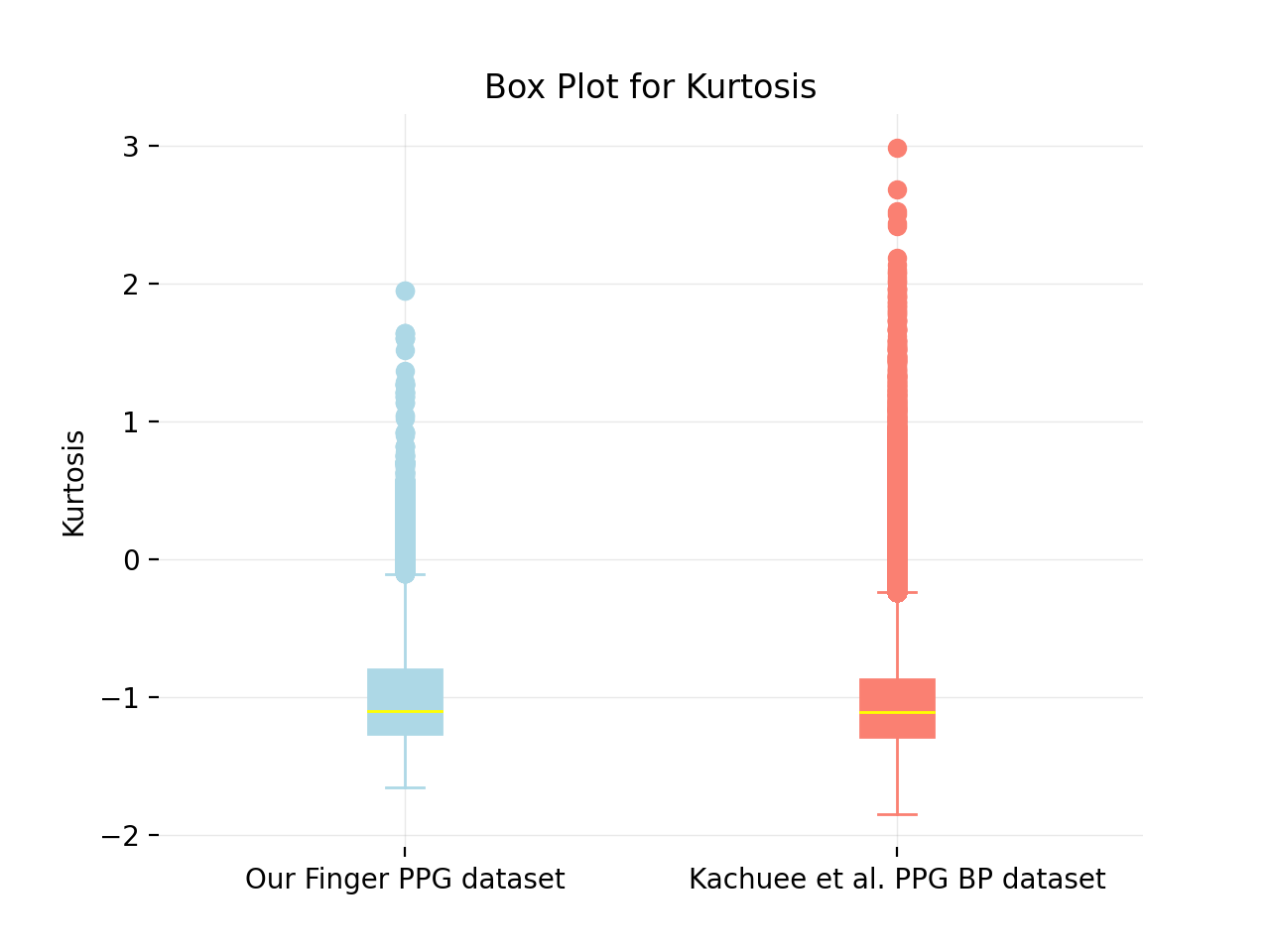}}
    \label{fig:sqi}
\end{figure}

\subsection{Data Analysis}

We acquired rPPG signals for BP estimation using a Canon camera to capture 1080P high-definition videos within a clinical setting. To enhance computational efficiency while preserving signal integrity, these videos were subsequently downsampled to a resolution of 72x72 pixels, followed by a  spatial averaging process applied independently to the Red, Green, and Blue channels of the selected regions. The Green channel was exclusively utilized as a proxy for the rPPG signal, due to its ability to retain all morphological waveform features crucial for accurate BP estimation, providing an advantage over more complex deep learning methods.

The rPPG signal underwent a template matching procedure post-acquisition aimed at isolating high-quality cardiac cycles. This process involved the computation of an averaged template across a patient’s entire signal, followed by the application of a correlation function to extract the most pertinent beat segments for subsequent analyses. This method ensures the selection of beats with maximal relevance and consistency, laying a solid foundation for the accuracy of the analysis.

For the purpose of BP estimation, the selected beats were introduced into a bespoke BP model architecture, comprising two distinct branches. The rPPG branch incorporated a convolutional neural network (CNN) consisting of five 1D convolutional layers, each succeeded by batch normalization and ReLU activation functions, to facilitate the extraction of localized PPG features. The output, a feature map of dimensions 128x32, was processed by a vision transformer, culminating in a comprehensive PPG embedding. Concurrently, the clinical branch of the model engaged a multilayer perceptron (MLP) to analyze a patient’s demographic and clinical attributes, amounting to 37 variables in total, generating a patient-specific embedding. The integration of PPG and patient embeddings was achieved through a fusion layer, followed by an additional MLP to conduct the final BP estimation. This architecture embodies a synergistic approach, leveraging both signal-derived and patient-centric data to enhance the precision and reliability of BP predictions. 



\section{Methods}

\subsection{Subjects and Data Collection}

The study protocol was approved by the University of Washington Institutional Review Board. 134 subjects were recruited from a general cardiology clinic between March and December 2022, of which 9 subjects had return visits, for a total of 143 patient visits. Written consent was obtained from each participant. Data collection was performed after routine clinical care to minimize disruption and to ensure ample time seated at rest (>5 minutes). Simultaneous rPPG, contact PPG, single-lead ECG, respiratory rate, BP, pulse rate, and blood oxygen saturation were obtained by the following method: Two exam rooms with consistent artificial lighting were used throughout the study. A digital single lens reflex (DSLR) camera (Canon EOS R5 Mirrorless Digital Camera with Canon EF 50mm f/1.8 STM Lens) recorded two consecutive two-minute videos of each subject's face to acquire the rPPG signal at 60 frames per second (fps) at manual zoom. BP was measured by an automated oscillometric cuff (Welch Allyn Connex Vital Signs Monitor 6000 Series)~\cite{jones2001validation} placed on the left upper arm while the patient was seated. The sequence for BP measurement and rPPG acquisition was as follows: the first BP measurement was immediately followed by the first two-minute video recording; then the second BP measurement was done and immediately followed by the second two-minute video recording; and then a third BP measurement was done. The first and second BP measurements were averaged to determine the BP for the first session; the second and third BP measurements were averaged to determine the BP for the second session. From start to end: continuous single-lead ECG was obtained from bilateral wrist leads. An optical sensor using green LED recorded PPG from a finger on the left hand. A standard pulse oximeter (Nellcor Portable SpO2 Patient Monitoring System) recorded blood oxygen saturation and pulse rate from another finger of the left hand. A thoracic band measured respiratory rate. The data from ECG, PPG, and thoracic band sensors were synchronously collected (ProComp Infiniti System with BioGraph Infiniti Software - T7500M). 

\subsection{Signal Processing}




The raw rPPG signals were extracted by spatially averaging the green channel pixels in each frame of the facial videos to produce a time-series rPPG signal \cite{verkruysse2008remote}. Prior work has shown that the green channel contains the highest PPG signal-to-noise ratio. By extracting the signal using a simple averaging technique, we aimed to preserve the PPG waveform morphology. A region around the eyes was excluded to reduce noise from eye motion (see Fig.~\ref{fig:architecture}). 

Subsequent preprocessing of the raw PPG and rPPG signals involved the application of a second-order bandpass Butterworth filter, with cutoff frequencies designated at 0.7 and 16 Hz, followed by min-max normalization to calibrate the amplitude of the filtered signals. Additionally, a peak detection algorithm facilitated pulse segmentation by pinpointing the systolic foot of each pulse wave, utilizing the scipy peak detection function for this purpose.

Given the inherent variability in rPPG and PPG quality across our experiments, not all extracted pulses met the stringent criteria required for accurate BP estimation. To ensure the reliability of our analysis, we implemented a rigorous screening process aimed at selecting only those signals that offer the highest fidelity for physiological interpretation. The initial phase of data screening excluded sessions that yielded fewer than 50 detectable peaks, adopting criteria that a peak height of 0.7 and a minimum inter-peak distance of 20, corresponding to one-third of a second at a frame rate of 60 fps. This led to the exclusion of 14 rPPG sessions (4.9\%) and 7 PPG sessions (2.5\%) from further analysis.

To assess the integrity of the captured pulse, we employed a cross-correlation technique between each beat and a template pulse derived from the session's average, thereby computing a Signal Quality Index (SQI) for each pulse. The SQI, ranging from 0 to 1, was used as a metric to evaluate waveform quality. Sessions in which the average SQI of five consecutive pulses dropped below 0.8 were considered to be below the necessary quality standard, resulting in the exclusion of an additional 28 rPPG sessions (9.8\%) and 33 PPG sessions (11.6\%). 

For the selection of waveform samples for BP estimation analysis, the study prioritized up to 20 five-beat windows based on their SQI rankings (SQI > 0.8). For rhythm classification experiments, we opted to utilize the complete signal data from each session to ensure a comprehensive representation of arrhythmia, thus omitting the SQI-based exclusion step. 










\subsection{Rhythm Classification}

Physician investigators classified the single-lead ECGs from all original sessions (n = 286) into four categories: normal sinus rhythm (n = 215), atrial fibrillation/flutter (n = 33), paced rhythm (n = 20), and frequent ectopy (n = 10). Eight sessions were excluded due to three missing and five indeterminate ECGs. After accounting for sessions dropped from signal processing and rhythm classification, the final numbers for each category are presented in Table \ref{tab:sys_bp_results_table}. Frequent ectopy was characterized as either frequent premature ventricular contractions (PVCs) or frequent premature atrial contractions (PACs), with frequent PVCs defined as more than 1 per minute in accordance with society guidelines derived from the Lown grading system~\cite{2017guidelinePVC}. Although not defined by a consensus standard, frequent PACs were also defined as more than 1 per minute, based on prior studies and for consistency~\cite{PACs}.

\subsection{Model Architecture}

The overall architecture of the network is illustrated in Figure~\ref{fig:architecture}. The network accepts 5 consecutive pulses of PPG or rPPG signal as input, along with optional supplementary information such as subject demographics, past medical history, and medications, to output systolic and diastolic BP. Initially, a 1-dimensional convolutional neural network (1D CNN) processes the input PPG/rPPG signal to extract low-level features, utilizing 1D convolutions with kernel sizes of 7, 5, and 3. The 1D CNN further incorporates average pooling layers and batch normalization. The latent output from the 1D CNN is fed into a 1-dimensional vision transformer (1D ViT)~\cite{dosovitskiy2020image} designed to discern the relationships between successive pulses. This 1D ViT features 8 heads and 512 feed-forward dimensions. The process concludes with fully connected layers that integrate the additional information with the PPG/rPPG features to estimate BP.

Following the exclusion criteria outlined in Signal Processing, the final number of sessions used to train each model is shown in Table~\ref{tab:sys_bp_results_table}.


\end{document}